\documentclass[12pt]{elsarticle}

\usepackage{graphicx}
\usepackage{pgf}
\usepackage{amssymb}
\usepackage{lineno}

\usepackage[numbers]{natbib}

\journal{Special Edition of Behavioural Brain Research}

\begin{document}

\begin{frontmatter}


\title{Adaptive coordination of working-memory and reinforcement learning in non-human primates performing a trial-and-error problem solving task}

\author{Guillaume Viejo\corref{cor1}\fnref{label1,label2}}
\ead{guillaume.viejo@mcgill.ca}

\author[label1]{Beno\^it Girard}

\author[label3]{Emmanuel Procyk}

\author[label1]{Mehdi Khamassi}

\address[label1]{Sorbonne Universit\'es, UPMC Univ Paris 06, CNRS, Institute of Intelligent Systems and
Robotics (ISIR), F-75005 Paris, France}

\address[label2]{Montreal Neurological Institute and Hospital,
3801 University Street,
Montreal, Quebec, Canada,
}
\address[label3]{University of Lyon, Universit\'e Claude Bernard Lyon 1, INSERM, Stem Cell and Brain Research Institute U1208, Lyon, France}

\begin{abstract}
Accumulating evidence suggest that human behavior in trial-and-error learning tasks based on decisions between discrete actions may involve a combination of reinforcement learning (RL) and working-memory (WM). While the understanding of brain activity at stake in this type of tasks often involve the comparison with non-human primate neurophysiological results, it is not clear whether monkeys use similar combined RL and WM processes to solve these tasks. Here we analyzed the behavior of five monkeys with computational models combining RL and WM. Our model-based analysis approach enables to not only fit trial-by-trial choices but also transient slowdowns in reaction times, indicative of WM use. We found that the behavior of the five monkeys was better explained in terms of a combination of RL and WM despite inter-individual differences. The same coordination dynamics we used in a previous study in humans best explained the behavior of some monkeys while the behavior of others showed the opposite pattern, revealing a possible different dynamics of WM process. We further analyzed different variants of the tested models to open a discussion on how the long pretraining in these tasks may have favored particular coordination dynamics between RL and WM. This points towards either inter-species differences or protocol differences which could be further tested in humans. 
\end{abstract}


\end{frontmatter}

%\linenumbers

	\section{Introduction}
    
 	The use of computational models relying on the Reinforcement Learning (RL) theory \cite{sutton1998reinforcement} in decision-making tasks greatly contributed to a better understanding of dopamine reward signals in the brain \cite{diederen2017}, neural activities in other brain areas such as prefrontal cortex and basal ganglia \cite{odoherty2004,dayan2008,ito2011,seo2014neural}, as well as alterations of brain activity and behavior in different pathologies \cite{huys2016,palminteri2017}. The understanding of the brain mechanisms at stake has been facilitated by the replication of central results in rodents, humans and non-human primates, such as dopamine-related reward prediction error signals \cite{hikosaka2008,gan2010,palminteri2015}, action value encoding in the striatum \cite{samejima2005representation,ito2009,palminteri2017}, forgetting mechanisms of action values \cite{barraclough2004,ito2009,Khamassi2015,niv2015}, or even neural correlates of parallel model-based and model-free learning processes \cite{johnson2007,Glascher2010,kennerley2011}. This enabled a transfer of knowledge between species and a more global understanding of possible neural architectures for the coordination of learning and decision-making processes.

However, it is not clear whether human and non-human primates always use similar learning and decision-making strategies in these types of tasks. In particular, \citet{collins2012much} recently showed that while most computational studies attempt to explain all aspects of behavior in terms of RL, human behavior in these types of tasks involve a combination of model-free reinforcement learning (MFRL) and working-memory (WM) processes. The evolution of human subjects' choices in their task was better explained as a weighted contribution of MFRL and WM in the decision process. Does monkey behavior show similar properties?

To answer this question, we propose to use the same model-based analysis approach that we recently employed in a human instrumental learning task \cite{viejo2015modeling}. In this study, we treated the WM component as a deliberative model-based system and a q-learning algorithm \cite{watkins1989learning} as the MFRL system and tested different processes to dynamically determine the contribution of each system in the decision of each trial of a task. We moreover proposed a novel method to compare the ability of different models to not only fit subjects' choices but also the trial-by-trial evolution of their reaction times, hence providing a finer description of behavior.

We previously found with this computational method that humans adaptively combined MFRL and WM, spending more or less time searching in working memory depending on the uncertainty of the different trials \cite{viejo2015modeling}. Here we tested the same models, plus new variations of these models, on monkey behavioral data in a deterministic four forced-choice problem-solving task \cite{procyk2006modulation,Quilodran2008,Khamassi2015}. As for humans, we found that the behavior of the five monkeys was better explained in terms of a combination of MFRL and WM, rather than by one of these decision systems alone. We nevertheless found strong inter-individual differences, some being better captured by the same coordination dynamics than for humans, others showing the opposite pattern. We further analyze different variants of the tested models to open a discussion on how the long pretraining in these tasks may have favored particular coordination dynamics between reinforcement learning and working memory. This points towards either inter-species differences or protocol differences which could be further tested in humans.

\section{Material and methods}
\label{S:2}

	\subsection{Problem solving task}
   
\begin{figure}[ht]
	\centering\includegraphics[width=1\linewidth]{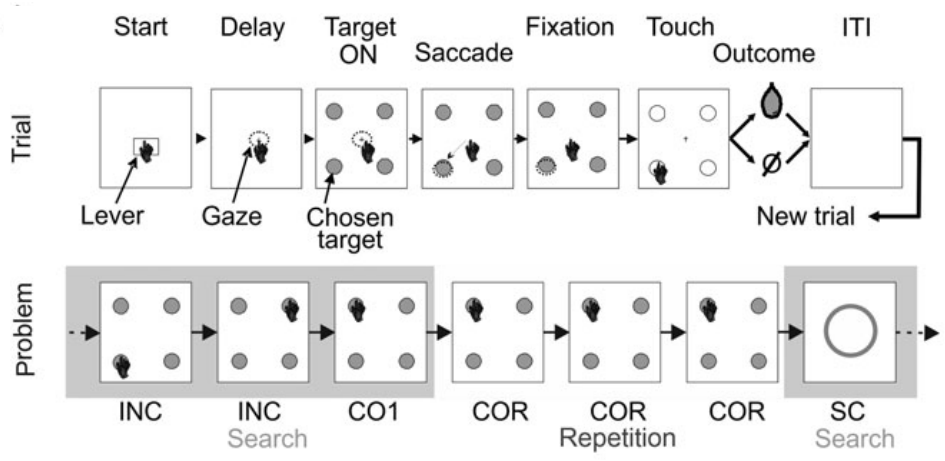}
	\caption{Trial-and-error Problem Solving Task. (Top) Successive events within a trial: the animal has to press a central lever and fixate his gaze on it until the trial starts; then targets appear, a go signal allows the animal to saccade towards its chosen target, then to touch the chosen target on the touch screen; finally, a reward is given or not, depending on the choice, and an inter-trial interval (ITI) is imposed. (Bottom) Successive trials within a problem. The animal performs a series of incorrect (INC) trials until finding the correct target and getting rewarded (first correct trial, CO1). Then the animal enters the repetition phase where it has to make at least three correct trials by repeating the choice of the same target. Finally, a signal-to-chance (SC) is presented indicating that the correct target is likely ($P=0.9$) to change location.}
	\label{F:1}
\end{figure}

Five monkeys were trained to discover by trial and error the correct target out of four possible targets as shown in figure~\ref{F:1} \cite{Quilodran2008,Khamassi2015}. A typical problem starts with a search phase during which the animal does incorrect trials (INC) until he performs the first correct trial (CO1). Then, a repetition phase starts for a various number of trials (from 3 to 11) during which the animal should repeat the correct action. This varying number of repetitions prevents the animal from anticipating the end of the problem. In the following analysis, only the first three repetition trials were compared with the models since they constitute the maximal common denominator between all achieved problems. At the end of the repetition phase, a signal indicates the beginning of a new problem. In 90\% of new problems, the rewarding target is different from the previous one. The successive events of a trial and a problem are represented in figure~\ref{F:1}.

	\subsection{Theoretical models}

	Our hypothesis is that monkeys solve this task through a combination of working-memory (WM) and model-free reinforcement learning (MFRL), as humans appear to do in similar tasks \cite{collins2012much,viejo2015modeling}. To test this hypothesis, we compared two different models representing different ways of coordinating WM and MFRL \cite{collins2012much,viejo2015modeling}, as well as models using either WM alone or MFRL alone to verify that they are not sufficient to explain the experimental data. We thus tested four computational models. Moreover, we tried different variations of these models to assess which particular computational mechanisms appear critical to explain the data.

\begin{figure}[ht]
	\centering\includegraphics[width=\linewidth]{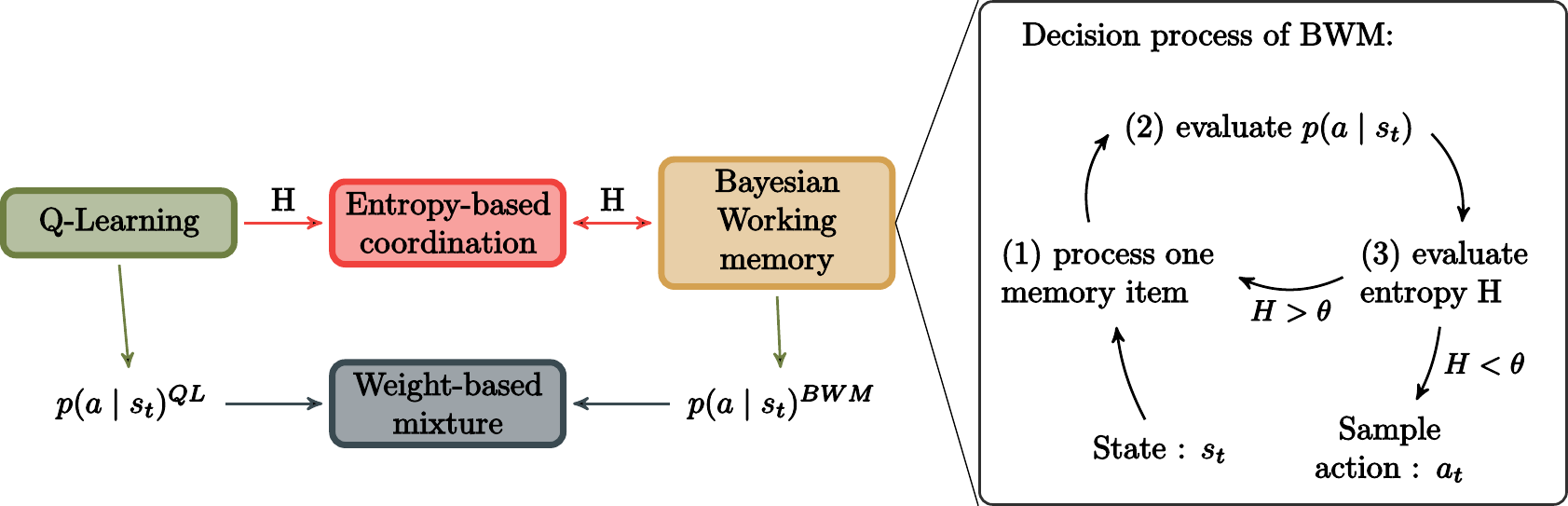}
	\caption{For all trials, the agent chooses an action with only one model (out of four). The deliberative behavior is represented by the Bayesian working memory model and the habitual behavior is represented by the q-learning. Between the two, the different models for interaction: Entropy-based coordination \citep{viejo2015modeling} and Weight-based mixture \citep{viejo2015modeling,collins2012much}. The right panel shows the decision process of the working memory. In the first step, the probability distribution $p(a,r|t_{0\rightarrow}i)$ is computed. The second step evaluates action probabilities and the third step compute the entropy H from the action probabilities. The goal of the cycle is to reduce iteratively the entropy H.}
	\label{F:2}
\end{figure}

With the exception of the q-learning algorithm \cite{watkins1989learning}, the models were first described in \citet{viejo2015modeling} without the new variations presented here. The task is modeled as one state (for each trial, we only model the decision moment where the animal chooses between the four targets, and gets a feedback for this choice), four actions (one per target) and two possible values of rewards (1 for a correct trial, 0 for an error). The four models and their relations are described in figure~\ref{F:2}.

		\subsubsection{Q-Learning (MFRL)}
        
The q-learning model is a standard model-free algorithm in the reinforcement learning field \cite{watkins1989learning}. It stores a table of q-values $Q(s,a)$ from which an action can quickly be drawn. After a transition to a new state s', the q-values are updated according to : 
\begin{equation}
	Q(s,a) \leftarrow Q(s,a) + \alpha [ r + \gamma max_a' Q(s', a') - Q(s,a) ]
\end{equation}
The sampling of an action is made through a soft-max equation : 
\begin{equation}
	p(a|s) = \frac{e^{\beta Q(s,a)}}{e^{\sum_{b\in A} \beta Q(s,b)}}
\end{equation}
At the initialization, the q-values are set to 0.0. 
		\subsubsection{Bayesian Working memory (BWM)}

The working memory model that we used chronologically stores the description (or memory item) of the events that occurred during past trials (i.e. chosen target, outcome). The number of memory items that can be stored is limited by a parameter $N$ optimized for each monkey. Each time a new item enters memory, the item older than $N$ trials is removed and memory decay is modeled by the convolution of the other items with a uniform distribution.

An element in memory contains the probability $p(a|t)$ of having performed an action in a certain past trial $t$ and the probability $p(r|a,t)$ of having observed a certain reward $r$ given an action $a$ in a past trial $t$. During the decision process, those probability mass functions are first combined with Bayes rule, then summed (see \citet{viejo2015modeling} for a full description of the equations). The sum of $i$ memory items gives $p(a,r|t_{0\rightarrow i})$ which is then reduced to $p(a|r,t_{0\rightarrow i})$. The index $0\rightarrow i$ indicates the number of memory items processed sequentially from the most recent one with an index of 0 to an oldest one with an index $i$. In this task, there are only two possible outcomes $r\in{0,1}$ and only one action is rewarding. At the beginning of a problem, when only incorrect trials have yet been experienced, only non-rewarded actions have been stored and untried actions should thus be favored. On the contrary, the probability of choosing the only action associated with the reward should be maximal if already observed. This reasoning has been summarized in the following equation : 
\begin{equation}
	Q(a) = \frac{p(a|r=1,t_{0\rightarrow i})}{p(a|r=0,t_{0\rightarrow i})}
\end{equation}
A simple normalization process allows the calculation of $p(a)^{BWM}$.

While previous working-memory models \cite{collins2012much} process all items in memory each time memory is screened, here the main novelty consists in examining memory items one by one until the model is confident enough about which action to perform. This is modeled as a dynamical allocation of the number $t_{0\rightarrow i}$ of memory items retrieved. The decision of evaluating the next memory item is dependent of the Shannon information entropy computed over the probability of action :
\begin{equation}
	H = - \sum_{a} \Big( p(a|t_{0\rightarrow i} ) \times log_2 p(a|t_{0\rightarrow i} )\Big)
\end{equation}
If H is above a parameterizable threshold $\theta$, the agent considers that it does not have enough information to decide and should thus evaluate the next element. If H is below $\theta$, the model considers that enough information has been incorporated into the probability of actions in order to make a decision. If all memory items within the list have been screened, the model is forced to make a decision on which action to perform. 

The number of memory items that can be processed for a given trial depends heavily on the history of past trials. If the correct action has been made at the previous trial, the number of memory items retrieved will be very likely one. If the correct action has not yet been found, the number of memory items retrieved will most likely be large. This feature of the Bayesian working memory model ended up being the crucial aspect by which human subjects' reaction times (RT) could be explained in \citet{viejo2015modeling}. Similarly, here the equation that relates the number of retrieved memory items with the simulated reaction times (sRT) is:
\begin{equation}
	sRT(trial) = ( log_2 (i+1) )^{\sigma} + H(p(a))
\end{equation}
The free parameter $\sigma$ controls the proportion of the first part of the equation in sRT. This equation is used for all the other models (in the case of the q-learning, $log_2(i+1) = 0$, so that its reaction time only depends on the contrast between learned action values).

In \cite{viejo2015modeling} as well as in this study, we assumed that slower RT correspond to the use of working memory. The main justification comes from the literature studying the effects of working memory as part of cognitive control processes, which has emphasized that more cognitive control is reflected by an increase in RTs \cite{cohen2004systems}. Some studies have further studied more specifically the link between RT variations and the balance between working memory and model-free reinforcement learning. For instance, in \cite{brovelli2008understanding,Brovelli2011}, the authors studied a task with human subjects similar to the one presented in this draft. Subjects had to associate by trial-and-error each stimulus with one correct action out of 5. The authors showed that this deliberative process (i.e. remembering wrong actions to select untried actions in order to find the right one) was associated with slower reaction times. When modeling this task \cite{viejo2015modeling}, we previously found that the working memory model presented in this draft, fitted best the behavior (choices and reaction times) during this deliberative process. 

		\subsubsection{Weight-based mixture (MTB)}
	
    The weight-based mixture model \cite{collins2012much} constitutes the first solution we tested for combining the Bayesian working memory model with the q-learning according to:
\begin{equation}
	p(a) = (1 - w_t)p(a)^{QL} + w_t p(a)^{MTB}
\end{equation}
The weight $w_t$ evolves after each trial according to : 
\begin{equation}
w_{t+1} = \frac{p(r_t|a_t)^{MTB} w_{t} }{p(r_t|a_t)^{MTB} w_{t} + p(r_t|a_t)^{QL} (1- w_{t}) }
\end{equation}
with $p(r_t|a_t)^{MTB}$ and $p(r_t|a_t)^{QL}$ being the relative likelihood that the corresponding model brought the reward. Thus, the weight evolves toward the most reliable strategy. 

		\subsubsection{Entropy-based coordination}

	The entropy-based coordination, first proposed in \cite{viejo2015modeling}, constitutes the second coordination solution that we tested. It explores the possibility of a closer interaction between the Bayesian working memory and the q-learning algorithm by conditioning the retrieval by the quantity of information contained in the working memory and the q-learning. The first point is to differentiate the entropies $H^{QL}$ and $H^{BWM}$ that can be computed for each strategy. The entropy $H^{QL}$ is computed with a soft-max function. The second point is that $H^{BWM}$ (being equal to $H^{max} = log_2(|Actions|)$ with $|Actions|$ the number of actions at the beginning of a trial) evolves inside a trial (hence reflecting a long inference process within the trial), while $H^{QL}$ evolves between trials (hence reflecting a long learning process across trials). The two entropies are used to control the retrieval probability of the working memory with the following sigmoid equation:
\begin{equation}
p(retrieval|H^{BWM}, H^{QL}) = 1 - \frac{1}{1+\lambda_1(n-i)exp^{-\lambda_2(2H^{max} - H_{0\rightarrow i}^{BWM} - H^{QL})}}
\label{E:sig}
\end{equation}
with $n$ the number of memory items present in the working memory list, $i$ the number of memory items already retrieved and $\lambda_1, \lambda_2$ gain parameters. This model thus discards the hard decision threshold $\theta$, used when the BWM operates in isolation. If the decision process is engaged, the q-values of each strategies are simply summed:
\begin{equation}
Q(a) = Q(a)_{0\rightarrow i}^{BWM} + Q(a)^{QL}
\end{equation} 

		\subsubsection{Variations of the models}
        
\begin{table}[htbp]
\centering
\begin{tabular}{l l l l l}
\hline 
& \textbf{BWM} & \textbf{Q-L} & \textbf{Mixture} & \textbf{Coordination} \\ 
\hline 
Variation 1  & original model & original model & original model & original model \\
& & & &  \\
Variation 2  & $\varnothing$ & $\gamma \in [0,1[$ & $\gamma \in [0,1[$ & $\gamma \in [0,1[$ \\
& & & &  \\
Variation 3  & $\varnothing$ & $\gamma \in [0,1[$ & $\gamma \in [0,1[$ & $\gamma \in [0,1[$  \\
& & $\neg$ INIT(Q-L) & $\neg$ INIT(Q-L) & $\neg$ INIT(Q-L) \\
& & & &  \\
Variation 4  & $\varnothing$ & $\gamma \in [0,1[$ & $\gamma \in [0,1[$ & $\gamma \in [0,1[$ \\
& & $\neg$ INIT(Q-L) & $\neg$ INIT(Q-L) & $\neg$ INIT(Q-L) \\
& & DECAY(Q-L) & DECAY(Q-L) &  DECAY(Q-L) \\	  
& & & &  \\
Variation 5 & ANT(BWM) & $\varnothing$ & $\gamma \in [0,1[$ & $\gamma \in [0,1[$ \\
&  & & $\neg$ INIT(Q-L) & $\neg$ INIT(Q-L) \\
&  & &  DECAY(Q-L) & DECAY(Q-L) \\
& & & ANT(BWM) & ANT(BWM) \\
& & & &  \\
Variation 6 & $\varnothing$ & $\varnothing$ & $\varnothing$ & $\gamma \in [0,1[$ \\
& & & & $\neg$ INIT(Q-L) \\
& & & & DECAY(Q-L) \\
& & & & META-L \\
& & & &  \\
Variation 7 & $\varnothing$ & $\varnothing$ & $\varnothing$ & $\gamma \in [0,1[$ \\
& & & $\neg$ INIT(Q-L) & $\neg$ INIT(Q-L) \\
& & & DECAY(Q-L) & DECAY(Q-L) \\
& & & THR($\delta$) & THR($\delta$) \\
& & & &  \\

\hline
\end{tabular}
\caption[Table of variations for the four possible models]{Variations of the Bayesian Working Memory model (BWM), the q-learning model (Q-L), the weighted mixture model and the entropy-based coordination model. The symbol $\varnothing$ designs the models that are not concerned by the tested variation. The variations with $\gamma \in [0,1[$ allows the discounting factor $\gamma$ to be optimized (set to 0 in the original version). With $\neg$ INIT(Q-L), the q-learning is not reinitialized at the beginning of a new block. The DECAY(QL) function allows the q-values to be forgotten given an optimized parameter of decay $\kappa$. With ANT(BWM), the working memory model evaluates the probability of action and the associated entropy during the outcome intervals in order to anticipate (thus gaining time) the decision for the next trial. This heuristic of anticipation is made only for the search phase. META-L is the meta-learning of average entropy value for each type of trial in order to bias the sigmoid function of the coordination model (see equation~\ref{eq:meta}). THR($\delta$) conditions the encoding of a past trial in the working memory based on the error prediction $\delta$ calculated from q-learning (see equation~\ref{eq:thrdela}). 
}
\label{tab:variations}				
\end{table}

The second novelty of the present paper is that from the original version of the four previously described models, we tested a number of variations (see table~\ref{tab:variations}). The aim is to examine which particular computational mechanisms of each model are critical to better capture the monkeys' behavior. The symbol $\varnothing$ indicates a model that is not concerned by the variation. 

In a first variation, the discount factor $\gamma \in [0,1[$ is optimized allowing the model to look in the future. In the original version of the q-learning, $\gamma = 0$ to account for the fact that transitions between states are randomized. Thus, there was no interest for the agent to learn the structure of states transition. In this framework, the task loops onto one state allowing the use of a discount factor. 

\citet{Khamassi2015} previously fitted various versions of the q-learning algorithm to this task. The most successful version considered that the information contained in the q-values were transmitted between problems i.e. the last rewarding action can bias the choice in a new search phase. This indicates some learning of the task structure by the monkeys: the rewarded target in the previous problem is very unlikely to be again rewarded in the present one. Therefore, we tested a second variation ($\neg$ INIT(Q-L)) where the q-learning values were not reinitialized. Nevertheless, it is likely that this strategy of non-reinitialization is more efficient if the values of q-learning are progressively erased along the task, as found in \cite{Khamassi2015}. Thus, we tested a third variation of q-learning (DECAY(Q-L)) with a decay factor to mimic progressive forgetting. The values are modified at each trial according to:
\begin{equation}
	Q(a) \leftarrow Q(a) + (1-\kappa)(Q_0 - Q(a)) 
\end{equation}
with $Q_0 = 0$ and $0 \leq \kappa \le 1$ the decay parameter.

For some cases (monkeys m and p, see figure~\ref{F:choice_rt_v1}), the reaction times show a specific pattern of diminishing during the search phase and re-increasing during the repetition phase, in contrast to what we previously observed in humans in a similar task \cite{viejo2015modeling}. This observation suggests that working memory is mostly involved during the repetition phase for these monkeys, since the use of working memory increases the reaction times. Given that the monkeys are trained for thousands of trials, it is very likely that the decision process during the search phase has been automatized. 
The first reason to hypothesize an automatization of choices during the search phase is the high number of trials that the monkeys did, leading the animals to learn an efficient search strategy. Their performance in the search and repetition phases is a good indicator of this fact. We hypothesize that the search strategy, as a rule, is a cognitive structure that has been learned and automatized such that it can be applied efficiently whatever the order of outcomes received during the search process. We have shown in previous publications that although the search is highly efficient, there is no fixed order in choices \cite{enel2016reservoir,procyk2006modulation}.The second reason is the dynamic of reaction times during the search phase : for the monkeys p, s and g, the reactions times are decreasing as the incorrect actions are revealed. Thus, we supposed that this dynamics was best translated by an automatic process. Yet, we did not found a fix individual idiosyncratic order of actions during the search phase.
However, monkeys must operate a deliberative process in order to avoid doing repetitive errors which have a high opportunity cost since making a correct response is very likely during repetition. Thus, a possibility that we have explored is the anticipation of the action by the working memory (ANT(BWM)). During the update of the models by the reward and for the search phase only, a simple heuristic lets the working memory retrieve all the memory items (including the most recent one) in order to prepare the probability distribution of actions for the next trial. Then, the entropy $H(p(a))$ of q-values (being either working memory alone or the combination with q-learning) will be lower at the onset of a trial without the cognitive load that would normally come with the retrieval of previous incorrect actions. Since the animal cannot anticipate the end of a problem (and has no interest in doing so), this heuristic is not applied to the repetition phase. 

To account for the over-training of monkeys, we tested the long-range learning of meta-parameters that can bias a model. We incorporated this idea in a new version of the entropy-based coordination model (META-L) with the addition of average entropy variables $\hat{H}^{BWM}_{type}$ and $\hat{H}^{QL}_{type}$ for each type of trial. Types can be trials during the search phase or trials during the repetition phase. Thus, the model learns a table that maps for each trial type the corresponding average entropies of the q-learning and the Bayesian working memory learned during thousands of trials performing the same the task. To average the entropies, the model is first tested in normal condition in order to store the distinct entropies computed at each trial. Finally, the sigmoid equation~\ref{E:sig} becomes:

\begin{equation}
p(retrieval) = 1 - \frac{1}{1+\lambda_1(n-i)exp^{-\lambda_2(2H^{max} - H_{0\rightarrow i}^{BWM} - H^{QL} + \hat{H}^{BWM}_{type} - \hat{H}^{QL}_{type})}}
\label{eq:meta}
\end{equation}
The addition of $\hat{H}^{BWM}$ will force trials with average high uncertainty of the working memory to decide faster. If $\hat{H}^{BWM}$ is high, the exponent $2H^{max} - H_{0\rightarrow i}^{BWM} - H^{QL} + \hat{H}^{BWM}_{type} - \hat{H}^{QL}_{type}$ will increase and gives a lower $p(retrieval)$. On the contrary, the average entropy $\hat{H}^{QL}$ modifies the sigmoid equation in the same direction as $H^{QL}$. A low average uncertainty on the q-learning favors fast decision of the model and conversely. 

The last modification incorporates the update of the working memory depending on the value $\delta_t$ of the temporal difference of the q-learning. The relation between working memory and temporal difference has already been explored in various models of reinforcement learning \cite{todd2009learning,rougier2005prefrontal} to account for observations of the physiological effects of dopamine on the circuitry of the prefrontal cortex \cite{goldman1995cellular,pessiglione2005effect,cools2011inverted,floresco2001delay}. Thus, we tightened the relation between the Bayesian working memory model and the q-learning algorithm (THR($\delta$)) for both models of interaction. The action of adding a new element in the working memory list is conditioned by:
\begin{equation}
[BWM] \leftarrow [p(a_t),p(r_t|a_t)] \ if \ \delta_t < \xi_1 \ or \ \delta_t > \xi_2
\label{eq:thrdela}
\end{equation}
For recall, the temporal difference is computed according to:
\begin{equation}
\delta_t = r_t + \gamma max_{a} Q(a) - Q(a)
\end{equation}
with $r_t$ the reward. In a nutshell, a large prediction error (being it positive or negative) induces an encoding of the last trial by the working memory since the q-learning has not converged. On the contrary, a small prediction error indicates a converging q-learning which can avoid a costly working memory update. 

		\subsubsection{Parameters optimization}
        
	As in \cite{viejo2015modeling,Lesaint2014,Lienard2014}, the parameters optimization was made using the SFERES toolbox \cite{Mouret2010} that implements the standard NSGA-2 evolutionary algorithm. Each variation of each model was optimized separately for each monkey. The scores of a set of parameters are the ability to maximize the likelihood on every step (correct and incorrect trials) that the model do the same actions as the monkey and minimize the mean-square error on representative steps between the average monkey reaction times and the simulated reaction times. The representative steps are defined as the trials inside the search phase and the following three repetitive trials separated according to the length of the search phase. Only the 0 to 4 errors blocks were considered in order to compute the representative steps. 
    The SFERES toolbox outputs the set of parameters that maximizes both the fit to choice and the fit to reaction times under the form of a Pareto front (see figure~\ref{F:pareto_fronts}). For each considered computational model, the Pareto front shows the solutions (i.e. the parameter sets) which are either not dominated by any other solution on at least one dimension, or not dominated by any other solution on at least one weighted combination of dimensions. For instance, a solution which gives the best fit on reaction times will be part of the Pareto front for the considered model even if its fit to choices is not the best. Similarly, a solution which gives the best mean fit on choices and reaction times will be part of the Pareto front even if there exists other solutions which are better at fitting either choice only or reaction times only.
    
    The methodology for selecting the optimal solution in a multi-dimensional optimization then relies on the use of an aggregation function. Such a function allows for the combination of numerical values $x_1,...,x_m$ into one value to rank all possible solutions. We used the Chebyshev distance in order to aggregate the normalized fitness functions (i.e. fit to choices and fit to reaction time) in one single value \cite{Wierzbicki1986}. The aggregation function for each x solution is defined as:
\begin{equation}
t(x) = \max\limits_{i\in M}\lambda_i \frac{\alpha_i-x_i}{\alpha_i-\beta_i} + \epsilon \sum\limits_{i=1}^{m}\lambda_i \frac{\alpha_i-x_i}{\alpha_i-\beta_i}
\label{eq:tchebytchev}
\end{equation}
with $\lambda \in \mathbb{R}^{M}_{+}$ a weight vector in order to bias the ranking if one fitness function is more important than the other (in our case $\lambda = 0.5$ for both), $\alpha_i = sup_{x\in \mathbb{X}}$ the optimal point and $\beta_i = inf_{x\in \mathbb{X}} x_i$ the worst combination of scores (although called the Nadir). 

\section{Results}

	\subsection{Simple models of the interaction between working memory and reinforcement learning}
	        
    	\subsubsection{Fit to choices only}

\begin{figure}[htbp]
\centering\includegraphics[width=0.5\linewidth]{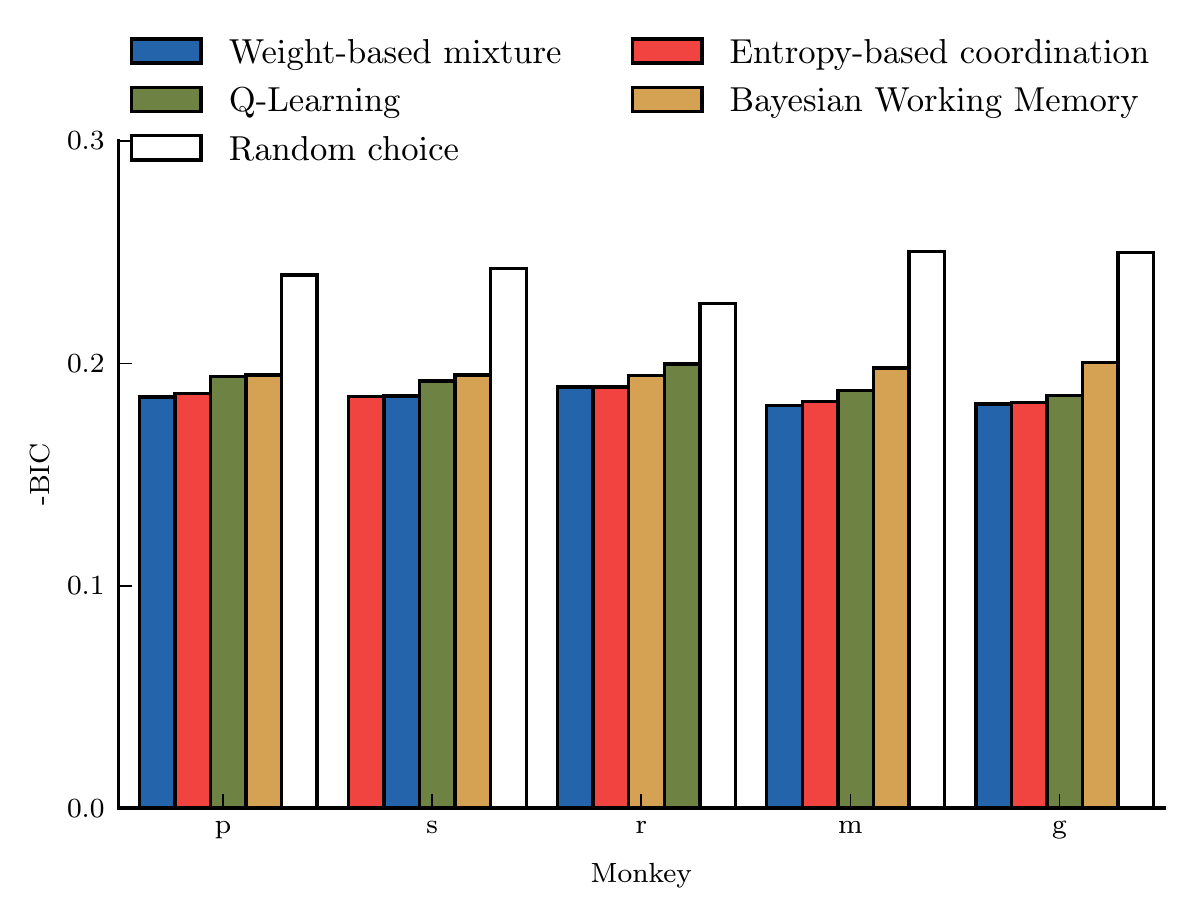}
\caption{Bayesian Information Criterion (BIC) score for each monkey and each model and compared to a random decision model (white bar). Overall, the dual models (entropy-based coordination or weight-based mixture) have a lower BIC score despite the fact that they have more free parameters (3 for q-learning, 4 for bayesian working memory, 7 for weight-based mixture and 8 for entropy-based coordination).}
\label{F:bic}
\end{figure}

We first evaluated the ability of each original model to replicate only the choices made by the monkeys. In figure~\ref{F:bic}, we compared the output of the optimization process using the Bayesian Information Criterion (BIC). Models were penalized proportionally to the number of free parameters. Nevertheless, the more complex dual models (entropy-based coordination or weight-based mixture) show better fits compared to simple models (Bayesian working memory or q-learning). Thus, this first analysis confirms that dual models made of the interaction between working memory and reinforcement learning are better than simple models at capturing the monkeys' behavior in this task. 

\begin{figure}[htbp]
\centering\includegraphics[width=1\linewidth]{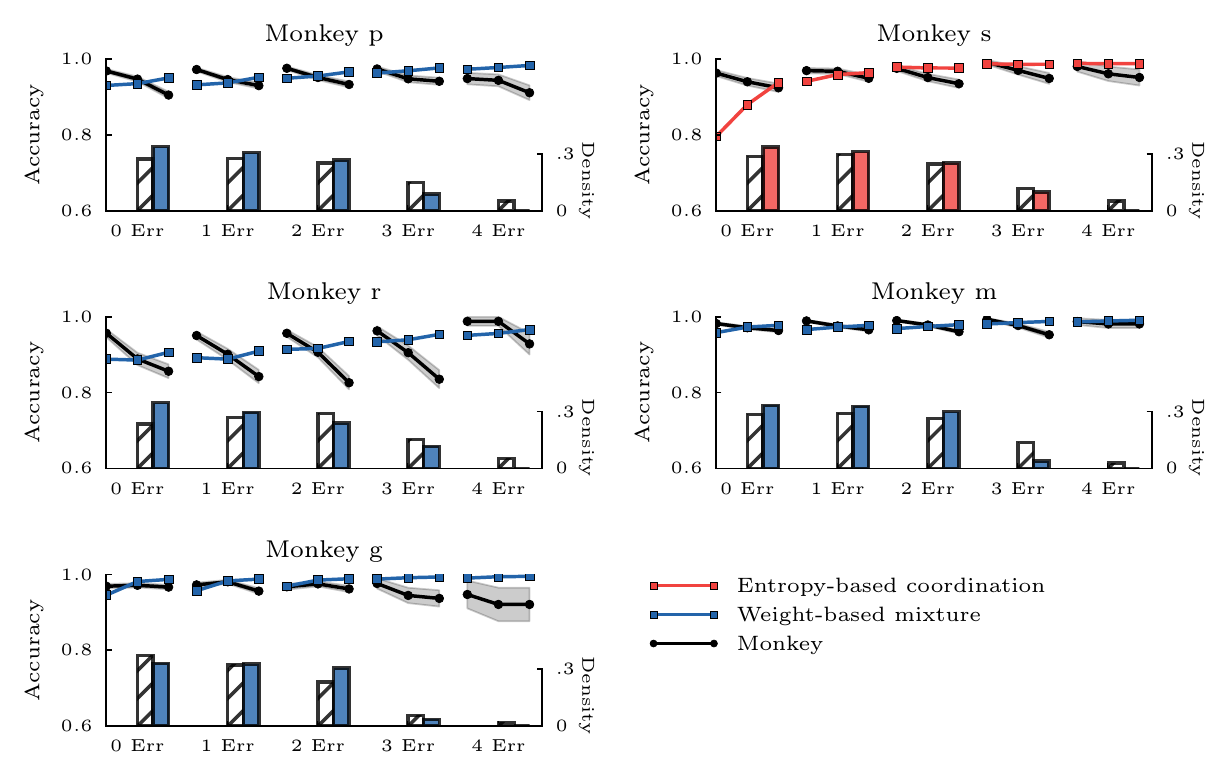}
\caption{Mean performances (mean $\pm$ sem) during the first three repetition trials for each monkey (black circles) and each best fitted model according to the BIC criterion (squares). Performances in repetition phase are averaged over problems with the same number of errors (from 0 to 4 errors). Each model is simulated 1000 times with the same chain of problems (same correct actions) than the corresponding monkey. Along the accuracy, the density of problem types (number of errors) is represented with bars for the monkey (dashed bar) and the model (full bar).}
\label{F:choice_only}
\end{figure}

\citet{palminteri2016} convincingly argued that model comparison without model simulation is not sufficient. In order to assess the validity of the fit according to BIC, we thus simulated the best model for each monkey. The set of best models is composed of the weight-based mixture models for 4 monkeys and the entropy-based coordination model for 1 monkey. During the simulation, only the list of problems (i.e. a list of indexes of the correct action) made by the monkey was used for transitioning between problems. The model was free to make its own choices and we repeated the experiment 1000 times. To display the performances, blocs (made of n trials of search phase plus 3 trials of repetition phase) were grouped for averaging according to the number of errors n-1 made during the search phase. For each group of blocs (defined by their number of errors n-1 during the search phase), the mean number of positive outcomes gained for each trial of the repetition phase (trials 1, 2 and 3 during which the animal should repeat the correct action) are then averaged to give the performances in repetition as shown in figure~\ref{F:choice_only}.

First, we found that the level of performance was roughly captured by most models. The performances of the monkeys were really high in repetition and all models reached this level of performance. Second, the striking observation is the inverse relation between the performance of the monkeys and the performance of the models. On average, monkeys have a decreasing performance between the first repetition and the third repetition trial. The models shows the opposite pattern with an increase of performance mostly due to the fact that dual models keep adding information about past trials in the working memory list. In figure~\ref{F:choice_only}, the bar density of each problem type (number of errors) is represented for each monkey and each best fitted model. We found that the problems count is slightly different between monkeys and models. When comparing the density of each type of problems, monkeys found the right action with less errors than the models (see density bars in figure~\ref{F:choice_only}).

These analyses suggest that an optimization of the model parameters in order to fit more aspects of monkeys' behavior, such as choices and reaction times, would be appropriate here.

		\subsubsection{Simulation of choices and reaction times}

\begin{figure}[htbp]
\centering\includegraphics[width=1.0\linewidth]{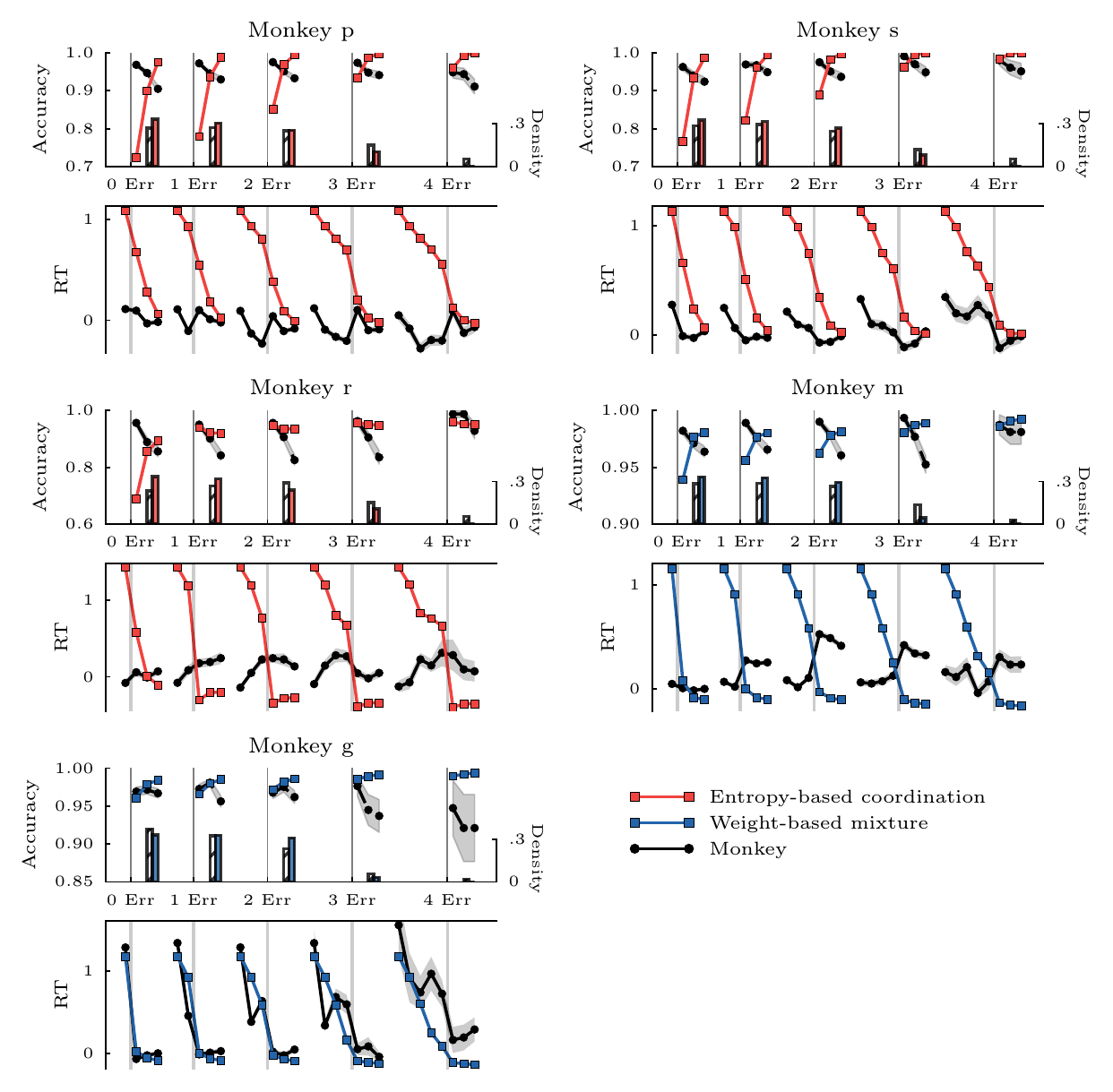}
\caption{For each monkey, the upper panel shows the mean performances (mean $\pm$ sem) and the lower panel shows the centered reaction times for the same trials. Vertical gray lines indicates the transition from search phase to repetition phase. Dotted lines shows the monkey's behavior and squared lines shows the best fitted model behavior. Similar to figure~\ref{F:choice_only}, performances and reaction times are averaged over problems with the same number of errors (from 0 to 4 errors). Each model is simulated 1000 times with the same chain of problems (same correct actions) than the corresponding monkey. Along the performance, the density of problem types (number of errors) is represented with bars for the monkey (dashed bar) and the model (full bar).}
\label{F:choice_rt_v1}
\end{figure}

In figure~\ref{F:choice_rt_v1}, we examined the fit of both choices and reaction times. Once again, only dual models were selected by the method of multi-criteria optimization tested in \citet{viejo2015modeling}. In addition to choices, we compared monkeys' reaction times with models' reaction times over representative steps. Reaction times are averaged over all problems of the same length (i.e. with the same number of errors). To the exception of monkey g, we found that the fit of reaction times was poorly performed by the original models (i.e. without the variations listed in table \ref{tab:variations}). Monkeys p and s showed progressively decreasing reaction times during the search phase, which were better fitted with the entropy-based coordination. Differently, monkey m showed constant reaction times during the search phase and a net increase during the repetition phase. Best fitted by the weight-based mixture, the model didn't manage to reproduce the global dynamics.

Overall, the original models of weight-based mixture and entropy-based coordination performed poorly in reproducing choices and reaction times in this trial-and-error search task with monkeys. In fact, the dynamics of the reaction times are much more diverse than the possibility that was given to those models for human behavioral data in \cite{viejo2015modeling}. In this original study, mean reaction times increased during the search phase and it was modeled by increments into the number of processed items within the working memory model in order to remember past incorrect actions (in order to avoid them). During the repetition phase, reaction times decreased and this was explained by the progressive shift to an habitual behavior (modeled using the q-learning) and thus a decrease in reaction times (since working memory was less used). In this task, the most common pattern is the opposite with decreasing reaction times during the search phase (monkeys p, s and g) and larger reaction times during repetition phase (monkeys m, p, s and g). To conclude, none of the fitted models so far displayed the ability to have first a decrease then an increase in reaction times. It is thus this ability that we tried to capture by testing more complex variations of the models. 

\subsection{Towards more complex models of interaction between working memory and reinforcement learning}

\begin{figure}[htbp]
\centering\includegraphics[width=0.8\textwidth]{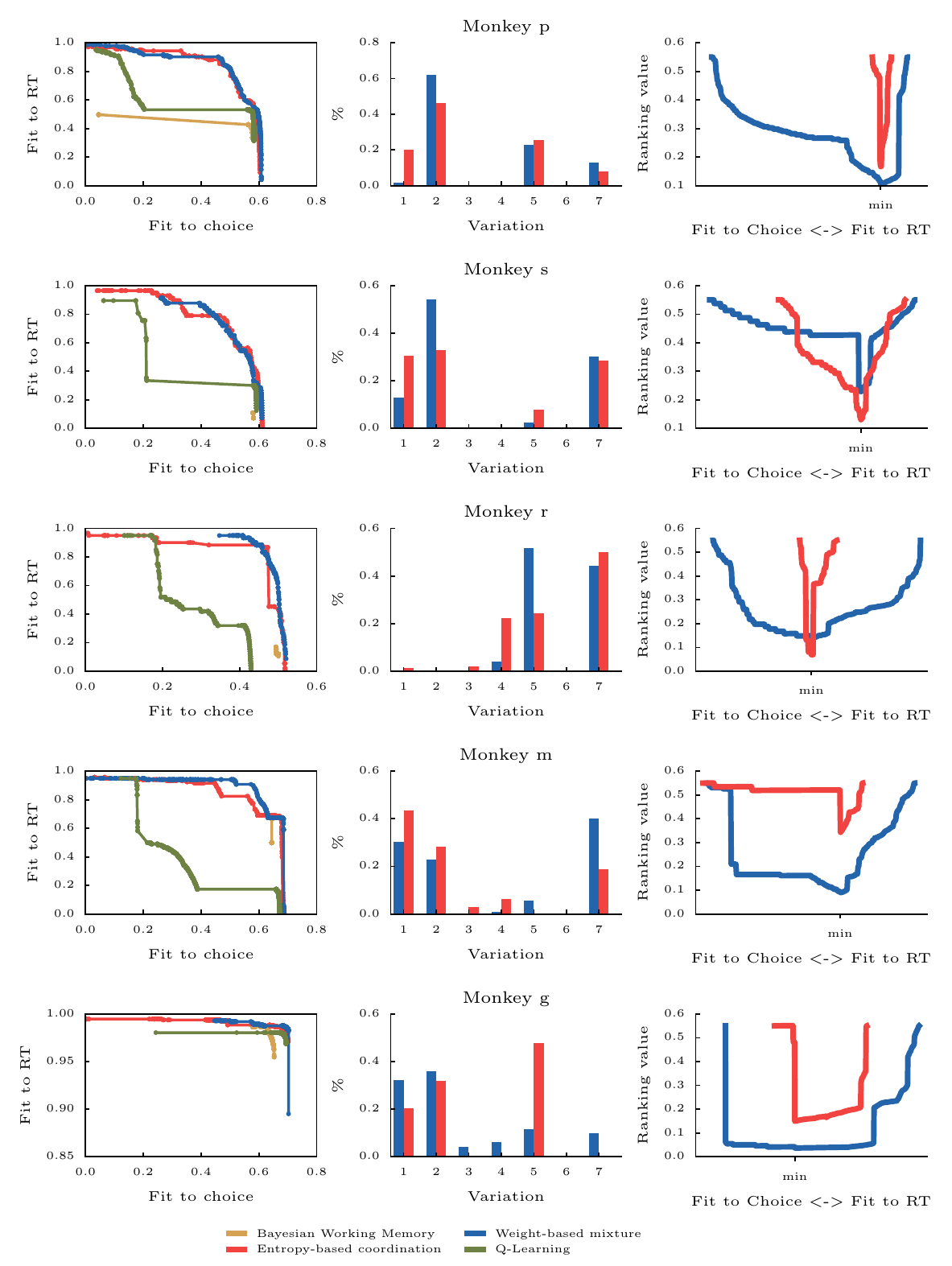}
\caption{Results of the optimization process for each monkey. The first column shows the pareto front for the weight-based mixture, the entropy-based coordination, the bayesian working memory and the q-learning. The second column shows the count for each variations within the weight-based mixture and entropy-based coordination model of parametersets selected within the pareto front. The third column displays the output of the Chebyshev aggregation function that converts the two dimension points of the Pareto fronts into a single value allowing points to be ranked.}
\label{F:pareto_fronts}
\end{figure}

	The results of the optimization process for the proposed variations of the initial model is shown in figure~\ref{F:pareto_fronts} for each monkey. The Pareto fronts (first column) show a domination of dual models once again confirming our first hypothesis that monkey behavior in this task can be better explained in terms of combination of working-memory and reinforcement learning processes rather than on each one alone. Moreover, for all monkeys we found that the Pareto fronts of weight-based mixture and entropy-based coordination were overlapping : no model took advantage over the other one. For the sake of clarity, the density of model variations inside each Pareto front only is represented (column 2). The last column shows the output of the Chebyshev function i.e. the aggregation of the fit to choice score and the fit to reaction times score for both the weight-based mixture and entropy-based coordination solutions. Since no dual models was definitively taking over, we decided to select and test the best solution according to the Chebyshev ranking (i.e. minimum value) for both models. For the entropy-based coordination, the best trade-off between fit to choice and fit to reaction times assigned the variation 5 ($\gamma \in [0,1[, \neg INIT(Q-L), DECAY(Q-L), ANT(BWM)$) to monkeys p, s and g and the variation 7 ($\gamma \in [0,1[, \neg INIT(Q-L), DECAY(Q-L), THR(\delta)$) to monkeys r and m.  For the weight-based mixture, the best trade-off assigned the variation 2 ($\gamma \in [0,1[$) to monkey g, the variation 5 to monkeys p and r and the variation 7 to monkeys m and s. As shown in figure~\ref{F:pareto_fronts}, those variations are effectively over-represented in the Pareto fronts of each monkey. 
    The overlaps between the two models suggest with higher confidence that the following computational mechanisms are important to explain monkeys' behavior: a positive discount factor ($\gamma \in [0,1[$) for all monkeys; forgetting without reset of action values ($\neg INIT(Q-L), DECAY(Q-L)$) for all monkeys except monkey g for which only the entropy-based coordination was selected with this mechanism; anticipation of the next trial during the search phase ($ANT(BWM)$) for monkey p (but sometimes also selected for monkeys g, r and s); modulation of storage in working-memory based on the sign and magnitude of reward prediction errors ($THR(\delta)$) for monkey m (but sometimes also selected for monkeys r and s). This confirms our prior use of action value forgetting mechanism for the subset of data associated with neurophysiological recordings in monkeys m and p \cite{Khamassi2015}. This nevertheless suggests a non-null discount factor in contrast to our prior work in both monkeys and humans in this type of tasks \cite{Khamassi2015,viejo2015modeling}, which will be further discussed later on.
    
	\subsubsection{Simulation of choices and reaction times}

\begin{figure}[htbp]
\centering\includegraphics[width=1.0\linewidth]{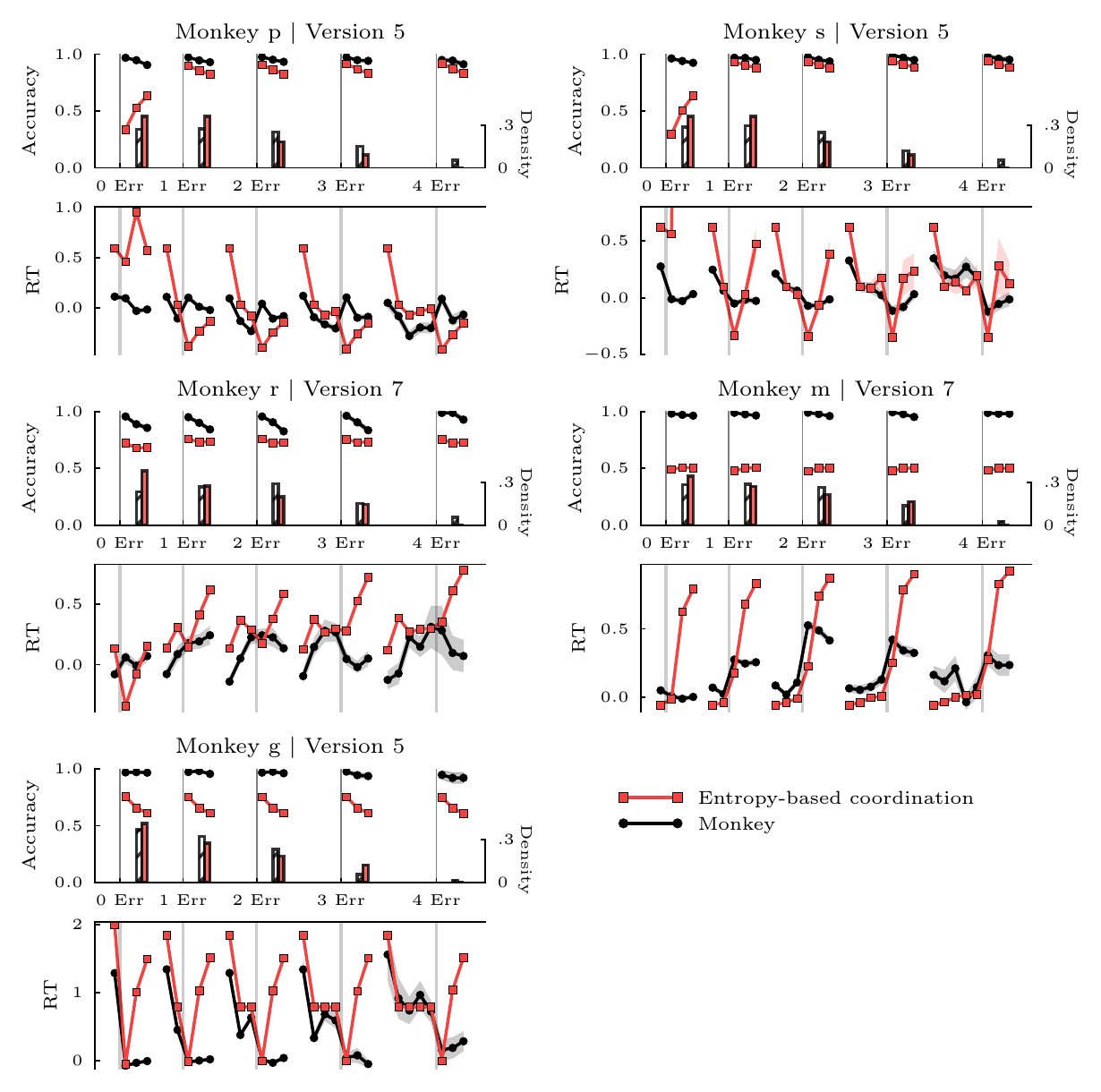}
\caption{Best simulated behavior for entropy-based coordination models. For each monkey, the upper panel shows the mean performances (mean $\pm$ sem) and the lower panel shows the centered reaction times (mean $\pm$ sem) for the same trials. Vertical gray lines indicates the transition from search phase to repetition phase. Dotted lines shows the monkey's behavior and squared lines shows the best fitted entropy-based coordination behavior. Versions were selected with a trade-off amongst solutions composing the Pareto front of this model as shown in figure~\ref{F:pareto_fronts}. The version for each model is displayed next to the monkey name. Each model is simulated 1000 times with the same chain of problems (same correct actions) than the corresponding monkey. Along the performance, the density of problem types (number of errors) is represented with bars for the monkey (dashed bar) and the model (full bar).}
\label{F:choice_rt_fusion}
\end{figure}

\begin{figure}[htbp]
\centering\includegraphics[width=1.0\linewidth]{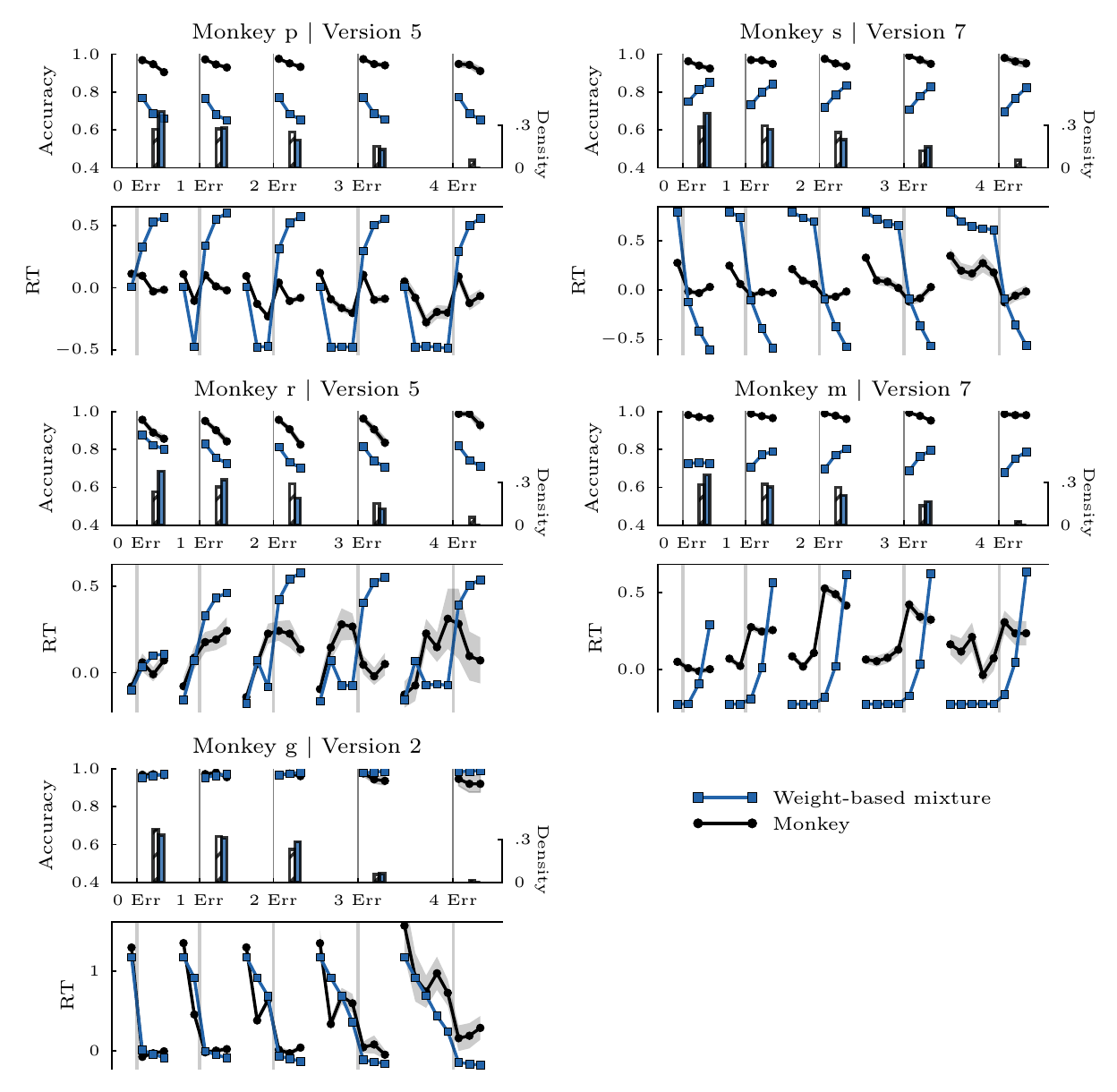}
\caption{Best simulated behavior for weight-based mixture models. For each monkey, the upper panel shows the mean performances (mean $\pm$ sem) and the lower panel shows the centered reaction times (mean $\pm$ sem) for the same trials. Vertical gray lines indicates the transition from search phase to repetition phase. Dotted lines shows the monkey's behavior and squared lines shows the best fitted weight-based mixture behavior. Versions were selected with a trade-off amongst solutions composing the Pareto front of this model as shown in figure~\ref{F:pareto_fronts}. The version for each model is displayed next to the monkey name. Each model is simulated 1000 times with the same chain of problems (same correct actions) than the corresponding monkey. Along the performance, the density of problem types (number of errors) is represented with bars for the monkey (dashed bar) and the model (full bar).}
\label{F:choice_rt_mixture}
\end{figure}

We tested each dual model's group of solutions as shown in figure~\ref{F:choice_rt_fusion} for entropy-based coordination and figure~\ref{F:choice_rt_mixture} for weight-based mixture. Overall, we found that the main caveat of the original models was corrected: reaction times could increase or decrease along the representative steps thus improving the fit to monkeys' reaction times. 

As usual when working on a multi-dimensional problem, the improvements on one dimension can lead to a degradation of the fit in another dimension. An instance of this issue is shown for the fit to choice translated into the performance of the simulated model in repetition. For all dual models (figure~\ref{F:choice_rt_fusion} and ~\ref{F:choice_rt_mixture}) to the exception of the monkey g with weight-based mixture, the performances in repetition were lower. For the reaction times, we observed improvements with our innovations. Contrary to the original models, our new models displayed various reaction times dynamics such as a decrease during the search phase and an increase during the repetition phase, in contrast to the pattern of reaction times that we previously found in humans \cite{viejo2015modeling}. 

	\subsubsection{Contribution of models}
          
\begin{figure}[htbp]
\centering\includegraphics[width=1.0\linewidth]{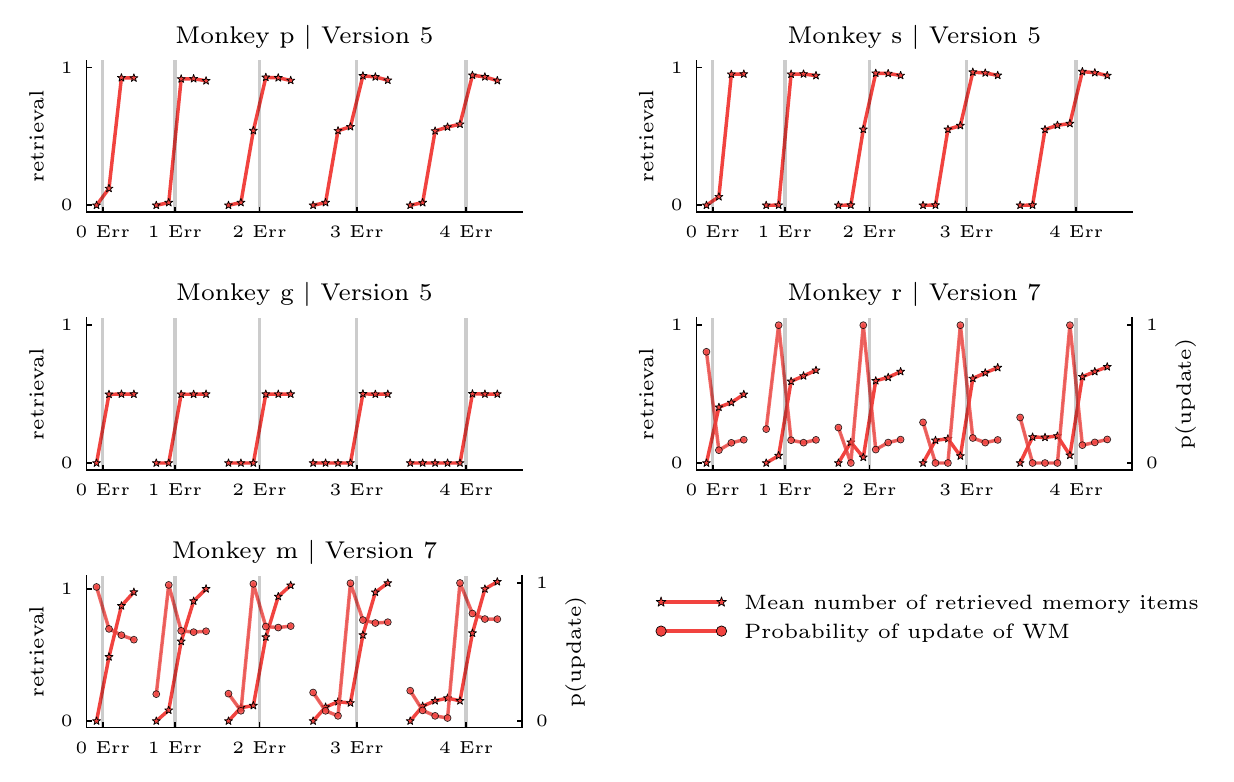}
\caption{Variables of entropy-based coordination model for each best fitted model. The star lines show the mean number of retrieved memory items. For monkeys best fitted with the version 7, the dotted lines show the probability of update of the working memory depending of the reward prediction error $\delta$ of the q-learning.}
\label{F:contribution_fusion}
\end{figure}

\begin{figure}[htbp]
\centering\includegraphics[width=1.0\linewidth]{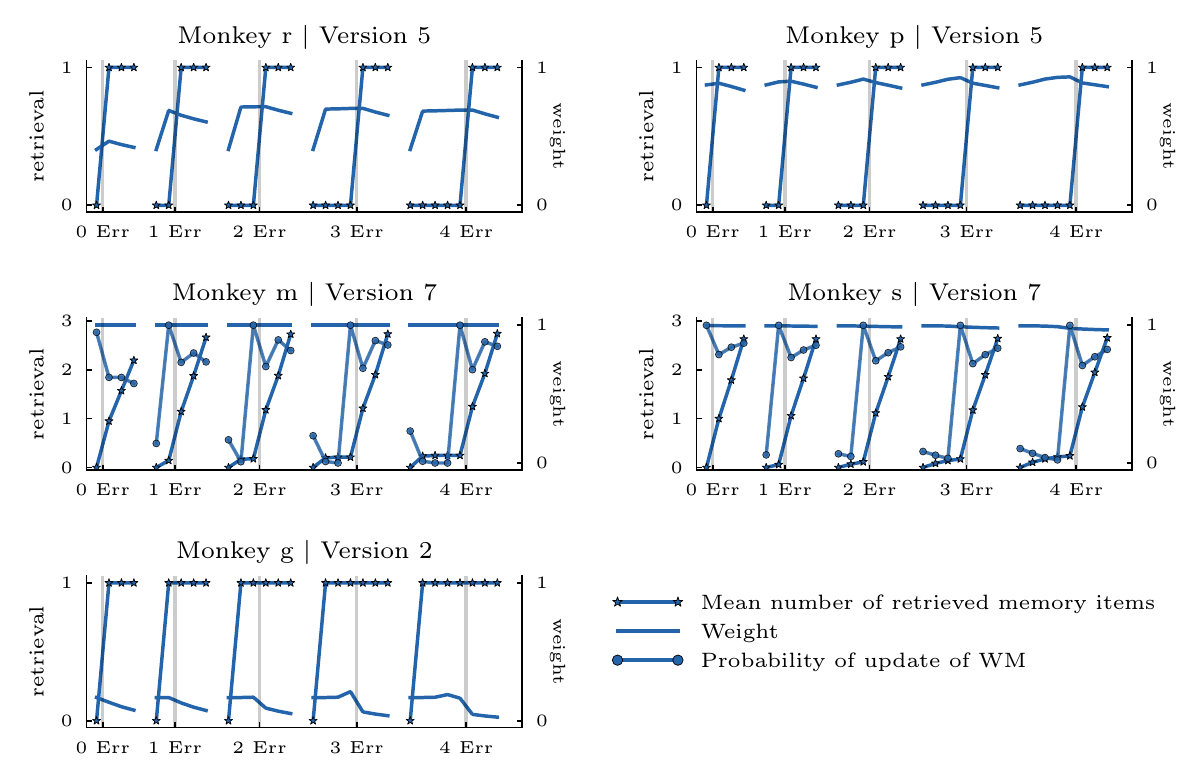}
\caption{Variables of weight-based mixture model for each best fitted model. The star lines show the mean number of retrieved memory items. The straight lines show the mean weight $w_t$. For monkeys best fitted with the version 7, the dotted lines show the probability of update of the working memory depending of the reward prediction error $\delta$ of the q-learning.}
\label{F:contribution_mixture}
\end{figure}

We then examined the internal dynamics of each best model in figure~\ref{F:contribution_fusion} for entropy-based coordination and figure~\ref{F:contribution_mixture} for weight-based mixture. While the relative contribution of Bayesian working memory and q-learning in weight-based mixture is easily measurable through the weight $w_t$, the relative contribution is less identifiable in entropy-based coordination. Thus, we decided to plot only the most relevant variable for each variation in order to decipher the dynamics of the model. 

For variation 5, the most relevant innovation is the anticipation of the next trial during the search phase. The whole content of working memory is recalled during the update phase after the agent receives a negative outcome during the search phase. This pre-retrieval modifies the probability of action for the Bayesian working memory (and by extension $H^{BWM}$) at the onset of a trial in the search phase which allows shorter reaction times. Thus, we looked at the number of retrieved memory items (star lines in figures~\ref{F:contribution_fusion} and~\ref{F:contribution_mixture}). For weight-based mixture, the relation was binary (monkeys r and p, figure~\ref{F:contribution_mixture}) : no memory items were retrieved right before the decision in the search phase and exactly one memory item (representing the last trial) was retrieved during the repetition phase. Besides, the mean weight $w_t$ for each representative step (normal lines in figure~\ref{F:contribution_mixture}) indicated the domination of the Bayesian working memory probability of action for the final decision. 
For the entropy-based coordination with the same variation, we found an intermediate level between no retrieval and constant retrieval. For monkeys p and s best explained by this model (figure~\ref{F:contribution_fusion}), the average number of retrieved items is 0.5 for the third to fifth trials within the search phase meaning that the model is more uncertain when anticipating the future trial. When items accumulate in the working memory list after 2 incorrect trials, the model has 50\% chance of retrieving the last item. Nevertheless, the entropy-based coordination is also able to produce the same pattern of binary retrieval during the search phase as shown by monkey g's fitted model. During the repetition phase, the model had the same behavior than the weight-based mixture with around 100\% chance of retrieving the last item. To summarize, we found here a possible explanation for the fast reaction times during the search phase (compared to the repetition phase) for monkeys p, s and m: the specific use of working-memory during repetition in order to ensure correct performance by anticipating the next trial. This constitutes an important prediction by the model whose neural correlates could be explored experimentally.

The second most used variation through dual models is variation 7 with the conditioning of the update of working memory by the reward prediction error $\delta_t$ computed from the q-learning. Within each panel with this variation in figure~\ref{F:contribution_fusion} and~\ref{F:contribution_mixture}, we computed the probability of update by averaging the number of times an item was integrated into the working memory list (dotted lines). For all fitted models using this new rule, we found that the probability of update was maximal at the end of the search phase i.e. when the first correct response is delivered to the agent. Thus, the fitted models of weight-based mixture (monkeys r and m) and entropy-based coordination (monkeys s and m) encoded positive outcomes inside the working memory only when the prediction errors were positive. In fact, the parameter $\xi_2$ that controls the upper bound was set between 0 and 0.5 for all fitted models (the parameter $\xi_2$ can take value in the range $[0,20]$ during the optimization process) and the reward prediction error $\delta=1$ at the first positive outcome. Strikingly, none of the fitted models had a lower bound  $\xi_1$ that allowed the encoding of a trial leading to a negative outcome. This makes sense since the retrieval of a memory item about a negative outcome brings less information and thus reduces less uncertainty than memory items about a positive outcome, as we previously explained in \cite{viejo2015modeling}, making the storage of memory items about negative outcomes less beneficial than positive ones. Because of this blocking of negative outcomes to enter working memory, the number of retrieved memory items is null during the search phase and only increases during the repetition phase for all models. Similarly to the previously described variation 5, this selectivity of encoding in working memory is what allowed the reaction times to be lower in the search phase compared to the repetition phase.  Lastly, the mean weight $w_t \approx 1.0$ in figure~\ref{F:contribution_mixture} for monkeys m and s indicates a reduction of the role of the q-learning to a simple prediction error signaler. 

Finally, monkey g was best fitted by the second variation of the weight-based mixture. The behavior, especially the reaction times, was already very well fitted with the original model. Here we found that the optimization of the q-learning with $\gamma \in [0,1[$ is the best fitted model according to the optimization process but the improvement of fit is minimal. Besides, the dynamic of the model shows a preference for the q-learning with a low $w_t$. Still, the working memory contributes to the final decision by constantly remembering the last trial. 

\section{Discussion}

	In this paper, we expanded and tested new models of coordination between working memory and reinforcement learning that were originally proposed in \citet{viejo2015modeling} in order to explain monkeys' behavior (choices and reaction times). During a succession of problems (defined by the correct action), monkeys had to find one correct target amongst four. When the correct target was found, animals repeated the correct action for a various number of trials (to prevent anticipation of the end of a problem). The first round of optimization with the original models \cite{viejo2015modeling} proved that a combination of working memory and reinforcement learning were better at explaining choices and reaction times than just working memory or reinforcement learning alone, which was the main hypothesis developed in this paper. 
    
	The hypothesis that distinct memory modules co-exist in the brain is supported by a range of lesion data in human \cite{scoville1957loss,corkin1968acquisition,corkin1984lasting,knowlton1996neostriatal}, in monkeys \cite{mishkin1978memory,squire1991medial} and in rodents \cite{sutherland1990hippocampus,mcdonald1993triple,packard1989differential}. Instrumental conditioning studies brought light on the interaction between distinct memory modules by deciphering the transfer of control that occurs between the early stage of learning (i.e when behavior is considered goal-directed) and late stage of learning (i.e when behavior is considered habitual). Lesion studies showed that different sets of brain areas supported those two stages of learning \cite{packard1996inactivation,coutureau2003inactivation,killcross2003coordination,yin2006role}. Thus, modeling studies used two different algorithms (respectively model-based learning algorithm and model-free reinforcement learning algorithm) for the two stage of learning with a transfer of control from model-based to model-free \cite{daw2005uncertainty,keramati2011speed}. Overall, the mapping between the reinforcement learning algorithm and brain activity during habitual behavior has been well described (see \cite{niv2009reinforcement}) . Evidences for the mapping between model-based learning algorithm and brain activity during goal-directed behavior are scarcer.  But evidence concerning specific neural substrates and properties for working-memory processes are supported by a vast literature (e.g. \cite{goldman1995cellular,esposito1995neural,ranganath2004inferior}. Here we do not assume a correspondence between working-memory processes and model-based learning. We simply consider that (1) WM belongs to a wider prefrontal cortex system dedicated to cognitive control, dedicated to inhibiting routine behaviors in response to environmental changes, and that (2) coordination mechanisms between model-based and model-free RL may be similar to coordination mechanisms between WM and model-free RL. In support of this, model-based and WM involve common prefrontal cortex regions \cite{stokes2013dynamic,balleine2010human}, regions such as the OFC being considered to encode the outcome of action and goal-directed action-outcome contingencies in working-memory \cite{frank2006anatomy}. Model-based processes actually do require working-memory when sequentially inferring the outcome of multiple actions within a cognitive graph \cite{wilson2014orbitofrontal}. Thus, there is a lot of possibilities in the combination and process of interaction of memory modules. The particular approach that we used to systematically compare different models of interaction between working memory and reinforcement learning using both choices and reaction times for each subject, is, in our sense, the best way to explore all the possibilities within the field of memory systems modeling.  	
    
	Originally developed to fit the behavior of humans in a visuo-motor association task \cite{brovelli2008understanding,Brovelli2011}, the models proved to be transferable to non-human primates. We then proceeded to improve the original models with different versions guided by the particular pattern of reaction times that we observed in monkeys: reaction times were lower for some trials during the search phase compared to the repetition phase. This observation opposes human's behavior for which the models were originally developed. The general hypothesis of the interaction process in humans was stated as followed: working memory is used during the search phase by remembering previous trials in order to avoid the selection of incorrect actions inducing an increase of reaction times as errors accumulate, and the q-learning gradually suppresses the use of working memory during the repetition phase as it converges toward the optimal decision with the accumulation of positive outcomes inducing faster reaction times. In sharp contrast, the general tendency of monkeys' reaction times was to accelerate during the search phase and to slightly slow down during the repetition phase. Thus, we made the hypothesis that working memory retrieval was not the main strategy that was used during the search phase or that it was used differently in combination with the reinforcement learning strategy. Oddly, the requirement of both strategies goes again the simplicity of the task : monkeys need to remember only the last correct action in order to succeed. While the task calls only for a working memory strategy, we found that a model-free reinforcement learning strategy was required. This result suggest that model-free might operate as a default strategy in the brain as previously proposed in \cite{Khamassi2015}. In \cite{seo2014neural}, the authors reported a similar dual model of working memory and model-free reinforcement learning strategies in order to explain the choices of monkeys confronted to a biased matching pennies game against a computer opponent. Similarly, monkeys complemented the model-free reinforcement learning algorithm with a more flexible strategies that was best reproduced with a working memory model. Nevertheless, they didn't explored the various combinations that a dual-strategy offers as in this study.

   In order to easily bias the use of working memory within a dual-strategy model, we first tested variations of q-learning with small changes: optimization of the discount factor, no initialization of q-values between problems and decay of q-values. In a second round of innovations, we tested more complex variations of working memory: (1) anticipation of the next trial during the search phase by preparing the probability of action, (2) meta-learning of mean entropies for suppressing the use of working memory when uncertainty is high on average, (3) encoding of past trials inside working memory conditioned by the reward prediction error from q-learning. Overall, we found that the anticipation of the next trial during the search phase and limiting the encoding of past trials were the best innovations to improve the fit to monkeys' behavior. 
   
   By analyzing the dynamics generated by simulating the best fitted models, we found that the anticipation of the next trial prevented in most dual models the retrieval process during the search phase and favored the retrieval of exactly one memory item (describing the last trial) during the repetition phase. The same process is at play when limiting the encoding of memory items (only the correct outcomes were included in the working memory list). Thus, it is very likely that the best theoretical model would incorporate the fact that working memory is somehow inhibited during the search phase and replaced by a more automatic behavior that can be different from the q-learning such as meta-learning of average uncertainty. We tested this approach by computing a table of mean entropies for each trial type and used this average uncertainties to bias the probability of retrieval of the entropy-based coordination models. Yet, this approach did not produce the best fit for explaining choices and reaction times. 
   
   This problem solving task has been studied in a series of articles using monkeys \cite{Khamassi2015,Quilodran2008,procyk2000anterior,rothe2011coordination} but also with humans in functional magnetic resonance imaging (fMRI) \cite{amiez2012modulation} and in electroencephalography (EEG) with 5 actions instead of 4 \cite{sallet2013modulation}. In both cases, the authors tried to correlate a reward prediction error $RPE = r_{obtained} - p_{correct}.r_{expected}$ with the cerebral activity recorded during the search phase. In fMRI, activations in the dorsal anterior cingulate cortex (midcingulate cortex), the frontal insular cortex, the striatum, the retrosplenial cortex and the middle dorsolateral prefrontal cortex correlated with a positive RPE. In other words : a high RPE means a lower expectation of reward and this is associated with a high cerebral activity. More interestingly, this correlation disappears for the negative RPE. In EEG, the authors analyzed the event-related potentials (ERP) when the subjects receive the outcome. Contrary to the results in fMRI, ERPs correlated with positive and negative RPE within the frontal regions. Besides, an ERP also appeared for the start signal of a new problem indicating a possible process of monitoring the structure of the task and not only positive and negative outcomes. Those experimental results tend to validate our second best fitted models, that condition the update of working memory by the reward prediction error $\delta$. Similar to the results in EEG, positive and negative reward prediction errors are used during the encoding phase of the task. Then, we found that only the positive outcomes were to be encoded within the working memory list which would resurged through neural activity (detected by fMRI) during the decision phase as a post-marker of the filtering during the update phase by the reward prediction error.  
   
      The opposite reaction times patterns that we observed here for the monkey behavioral data \cite{Khamassi2015} compared to the human data \cite{Brovelli2011} could be seen as an indication of inter-species differences in learning and decision-making strategies. When fitting our dual models to human data we previously found that working-memory was important to prevent the repetition of errors during search \cite{viejo2015modeling}. Here the model-based analyses suggest that working-memory in the five studied monkeys is important to ensure the repetition of correct response once the correct target has been discovered, while working-memory processes may present a cost-benefit ratio  too low during search (retrieving a memory item about a negative outcome is less informative because it only tells which target not to select, while a positive outcome directly tells which target to select). An alternative explanation to the inter-species differences may be that the long pretraining phase that monkeys underwent for this task may have enabled them to learn more aspects of the task structure and hence to restrict their use of working-memory. We have tried to capture this phenomenon in two different ways here: 1) using meta-learning where the model learns the average uncertainty that results from deliberative processes at experienced type of trials (this enables the model to automatically learn that using working-memory during the search phase does not reduce much the uncertainty), and 2) using trial anticipation where the model retrieves a memory item in preparation of the decision at the next trial to ensure correct response repetition. The meta-learning variation of the tested models was never selected as best model. Nevertheless, the trial anticipation variation was consistently selected as best model for monkey p, and sometimes also for monkeys g, r and s. An important experimental prediction of these model variations is that humans undergoing the same long pretraining phase for this task would be able to decipher the task structure and thus to show the same opposite reaction times patterns than those observed in humans without pretraining in the task of \citet{Brovelli2011}. Similarly, humans being given detailed instructions about the task could also extract sufficient information about task structure to display the same opposite reaction time patterns. A perspective of this work would be to apply the same model-based analysis to the human data gathered in the same problem-solving task by \citet{sallet2013modulation}. The specific design of this task may have favored particular coordination dynamics between reinforcement learning and working memory.
  
  The study of this problem solving task with non-human primates \cite{Khamassi2015,Quilodran2008,procyk2000anterior,rothe2011coordination} and human subjects \cite{amiez2012modulation,sallet2013modulation} clearly shows cerebral activity associated with the evaluation, encoding and monitoring of uncertainty associated with decisions. Simple models of coordination between working memory and reinforcement learning or working memory alone do not have this ability as they just encode the description of a trial. Meta-learning or anticipation as tested in this paper could thus bridge the gap between dual strategies models and high-level cognitive models as the fitted models indicate. To conclude, those models would be perfect to look for new computational variables that can be used for correlation with neuronal activity and to elucidate the processes taking place in the underlying brain structures.



%% References

\clearpage
\bibliographystyle{model1-num-names}

\bibliography{Biblio.bib}


\end{document}